\documentclass[runningheads]{llncs}
\usepackage{graphicx}
\usepackage{url}

\usepackage{epstopdf}
\usepackage{booktabs}
\usepackage{epsfig}
\usepackage{amsmath}
\usepackage{amssymb}
\usepackage{mathtools}
\usepackage{url}
\usepackage[noend]{algpseudocode}
\usepackage{algorithmicx,algorithm}
\usepackage{bbm}

\usepackage{makecell}
\usepackage{color}
\usepackage{url}
\usepackage{mathrsfs}
\usepackage{amsmath}
\usepackage{amssymb}
\usepackage{setspace}
\newcommand{\ie}{\textit{i.e.}, }
\newcommand{\eg}{\textit{e.g.}, }

\newcommand{\myPara}[1]{\vspace{0.02in}\noindent\textbf{#1}}
\graphicspath{{./figs/}}
\usepackage{multirow} 
\usepackage{makecell} 
\usepackage{bbding}
\usepackage{pifont}
\usepackage{color}
\usepackage[colorlinks,linkcolor=red]{hyperref}
\usepackage{lipsum}

\begin{document}

\title{USCL: Pretraining Deep Ultrasound Image Diagnosis Model through Video Contrastive Representation Learning}

\titlerunning{Ultrasound Contrastive Representation Learning}

\author{Yixiong Chen \inst{1,4,}\thanks{The first two authors contributed equally. This work was done at Shenzhen Research Institute of Big Data (SRIBD).} \and
Chunhui Zhang\inst{2,\star} \and
Li Liu\inst{3,4} \and Cheng Feng \inst{5,6}, Changfeng Dong \inst{5,6}, Yongfang Luo \inst{5,6}, Xiang Wan \inst{3,4}}


\authorrunning{Y. Chen et al.}

\institute{School of Data Science, Fudan university, Shanghai, China \and
Institute of Information Engineering, Chinese Academy of Sciences, Beijing, China \and
Shenzhen Research Institute of Big Data, Shenzhen, China \\
\email{liuli@cuhk.edu.cn}\\
\and The Chinese University of Hong Kong Shenzhen, Shenzhen, China \and Shenzhen Third People's Hospital, Shenzhen, China \and Southern University of Science and Technology, Shenzhen, China}

\maketitle              
\begin{abstract}

 Most deep neural networks (DNNs) based ultrasound (US) medical image analysis models use pretrained backbones (\eg ImageNet) for better model generalization. However, the domain gap between natural and medical images causes an inevitable performance bottleneck. To alleviate this problem, an US dataset named US-4 is constructed for direct pretraining on the same domain. It contains over 23,000 images from four US video sub-datasets. To learn robust features from US-4, we propose an US semi-supervised contrastive learning method, named USCL, for pretraining. In order to avoid high similarities between negative pairs as well as mine abundant visual features from limited US videos, USCL adopts a sample pair generation method to enrich the feature involved in a single step of contrastive optimization. Extensive experiments on several downstream tasks show the superiority of USCL pretraining against ImageNet pretraining and other state-of-the-art (SOTA) pretraining approaches. In particular, USCL pretrained backbone achieves fine-tuning accuracy of over 94\% on POCUS dataset, which is 10\% higher than 84\% of the ImageNet pretrained model. The source codes of this work are available at \url{https://github.com/983632847/USCL}.


\end{abstract}

\section{Introduction}
Due to the low cost and portability, ultrasound (US) is a widely used medical imaging technique, leading to the common application of US images~\cite{born2020pocovid,Somphone2014MICCAI} for clinical diagnosis. To date, deep neural networks (DNNs)~\cite{he2016deep} are one of the most popular automatic US image analysis techniques. When training DNN on US images, a big challenge is the data scarcity, which is often dealt with parameters transferred from pretrained backbones (\eg ImageNet~\cite{ILSVRC15} pretrained VGG or ResNet). 
But model performance on downstream tasks suffers severely from the domain gap between \textit{natural} and \textit{medical} images~\cite{ke2021chextransfer}. There is a lack of public well-pretrained models specifically for US images due to the insufficient labeled pretraining US data caused by the high cost of specialized annotations, inconsistent labeling criterion and data privacy issue.

\begin{figure}[t]
	\centering\centerline{\includegraphics[width=0.95\linewidth]{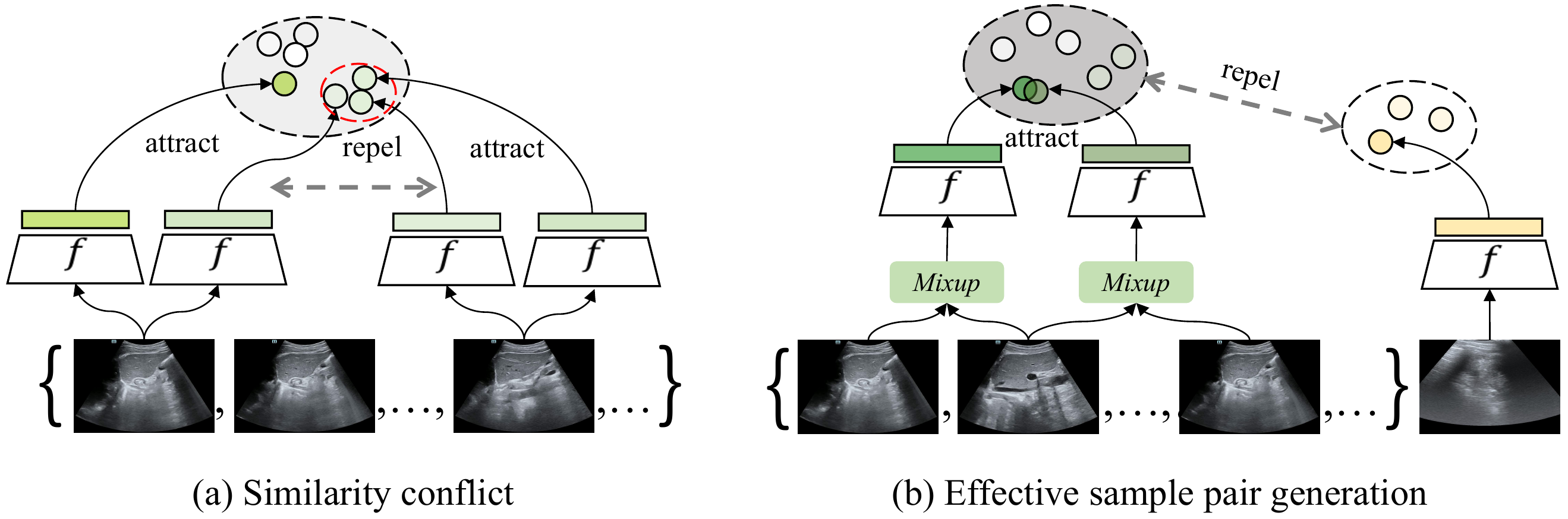}}
	\caption{Motivation of USCL. SPG tackles the harmful similarity conflict of traditional contrastive learning. (a) Similarity conflict: if a negative sample pair comes from different frames of the same video, they might be more similar than positive samples augmented from a frame, which confuses the training. (b) SPG ensures negative pairs coming from different videos, thus are dissimilar. The sample interpolation process also helps the positive pairs to have appropriate similarities, and enriches the feature involved in comparison. Representations are learned by gathering positive pairs close in representation space and pushing negative pairs apart.}
	\label{fig:Motivation}
\end{figure}

Recently, more and more literature tend to utilize unsupervised methods~\cite{celebi2016unsupervised,he2020momentum} to avoid medical data limitation for pretraining. The common practice is to pretrain models with pretext tasks and evaluate the representations on specific downstream tasks. 
Yet most existing methods can only outperform ImageNet pretraining with high-cost multi-modal data~\cite{jiao2020self2,li2020self}. To get powerful pretrained models from US videos, we first build an US video dataset to alleviate data shortage. 
Secondly, contrastive learning\cite{vu2021medaug,he2020momentum,chen2020simple} is also exploited to reduce the dependence on accurate annotations due to its good potential ability to learn robust visual representations without labels. 
However, given the fact that most US data are in video format, normal contrastive learning paradigm (\ie SimCLR~\cite{chen2020simple} and MoCo~\cite{he2020momentum}, which considers two samples augmented from each image as a positive pair, and samples from different images as negative pairs) will cause high similarities between negative pairs sampled from the same video and mislead the training. This problem is called \textit{similarity conflict} (Fig.~\ref{fig:Motivation} (a)) in this work. \textit{Thus, is there a method which can avoid similarity conflict of contrastive learning and train a robust DNN backbone with US videos?}

To answer this question, we find that image features from the same US video can be seen as a cluster in semantic space, while features from different videos come from different clusters. We design a sample pair generation (SPG) scheme to make contrastive learning fit the natural clustering characteristics of US video (Fig.~\ref{fig:Motivation} (b)). Two samples from the same video act as a positive pair and two samples from different videos are regarded as a negative pair. In this process, two positive samples can naturally be seen as close feature points in the representation space, while negative samples have enough semantic differences to avoid similarity conflict. In addition, SPG does not simply choose frames as samples (\eg key frame extraction~\cite{liu2003novel}), we put forward sample interpolation contrast to enrich features. Samples are generated from multiple-frame random interpolation so that richer features can be involved in positive-negative comparison. This method makes the semantic cohesion appear at the volume level of the ultrasound representation space~\cite{kwitt2013localizing} instead of the instance level. Combined with SPG, our work develops a semi-supervised contrastive learning method to train a generic model with US videos for downstream US image analysis. Here, the whole framework is called \textit{ultrasound contrastive learning (USCL)}, which combines supervised learning to learn category-level discriminative ability, and contrastive learning to enhance instance-level discriminative ability. 

\begin{table*}[t]
	{\small
	\renewcommand\arraystretch{0.90}
	\caption{Statistics of the {US-4} dataset containing 4 video-based sub-datasets. The total number of images is 23,231, uniformly sampled from 1051 videos. Most videos contain 10$\sim$50 similar images, which ensures the good property of semantic clusters.}
	\label{tab:US-4}
	\begin{center}
		\setlength{\tabcolsep}{0.9mm}{
			\begin{tabular}{c|c|c|c|c|c|c|c}
				\hline
				Sub-dataset & Organ &  Image size & Depth & Frame rate &Classes & Videos & Images \\ 
				\hline
				Butterfly~\cite{ButterflyData} & Lung & 658$\times$738 & - & 23Hz & 2 & 22 & 1533   \\
				CLUST~\cite{Somphone2014MICCAI} & Liver & 434$\times$530 & - &  19Hz & 5 & 63   & 3150    \\ 
				{Liver Fibrosis} & Liver & 600$\times$807 & $\sim$8cm & 28Hz & 5 & 296 & 11714    \\ 
				{COVID19-LUSMS} & Lung & 747$\times$747 & $\sim$10cm & 17Hz & 4 & 670 & 6834   \\ 
				\hline
			\end{tabular}
		}
	\end{center}
    }
\end{table*}


\section{US-4 Dataset}
\begin{figure}[t]   
	\centering\centerline{\includegraphics[width=1.0\linewidth]{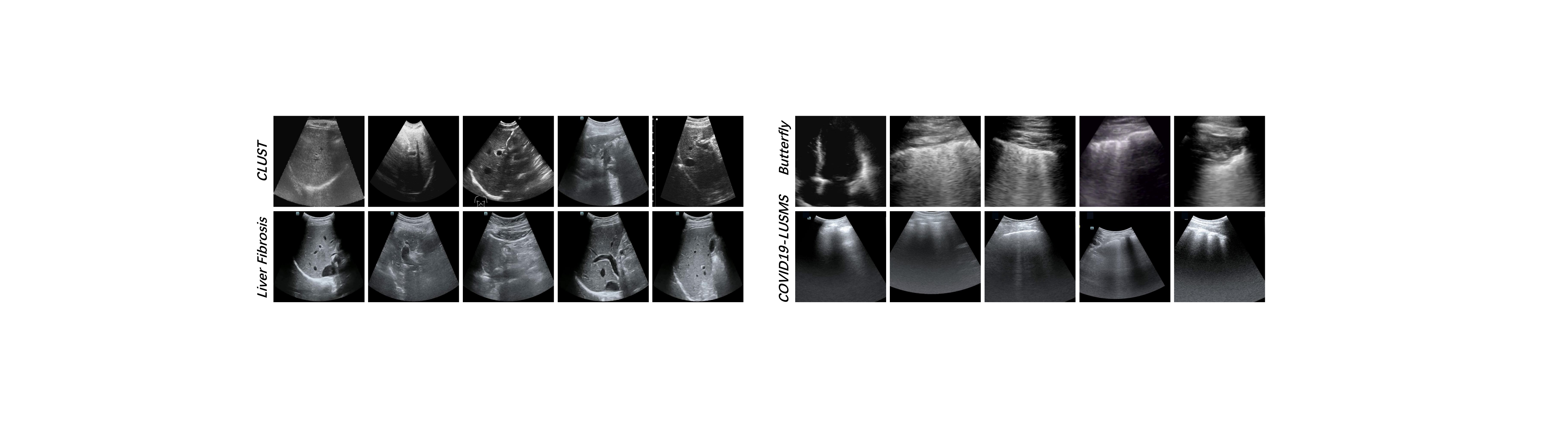}}
	\caption{Examples of US image in {US-4}.}
	\label{fig:Examples_US4}
\end{figure}

{\color{black}In this work, we construct a new US dataset named US-4, which is collected from four different convex probe~\cite{born2020pocovid} US datasets, involving two scan regions (\ie lung and liver). Among the four sub-datasets of US-4, \textit{Liver Fibrosis} and \textit{COVID19-LUSMS} datasets are collected by local sonographers~\cite{gao2021multi,liu2020semi}, \textit{Butterfly}~\cite{ButterflyData} and \textit{CLUST}~\cite{Somphone2014MICCAI} are two public sub-datasets. The first two sub-datasets are collected with \textit{Resona 7T} ultrasound system, the frequency is FH 5.0 and the pixel size is 0.101mm - 0.127mm. All sub-datasets contain labeled images captured from videos for classification task. In order to generate a diverse and sufficiently large dataset, images are selected from original videos with a suitable sampling interval. For each video with frame rate $T$, we extract $n=3$ samples per second with sampling interval $I\!=\! \frac{T}{n}$, which ensures that US-4 contains sufficient but not redundant information of videos. This results in 1051 videos and 23,231 images. The different attributes (\eg depth and frame rate) of dataset are described in Tab.~\ref{tab:US-4}. The US-4 dataset is relatively balanced in terms of images in each video, where most videos contain tens of US images. Some frame examples are shown in Fig.~\ref{fig:Examples_US4}.}

\section{Methodology}
\begin{figure}[t]
	\centering 
	\includegraphics[width=1.0\linewidth]{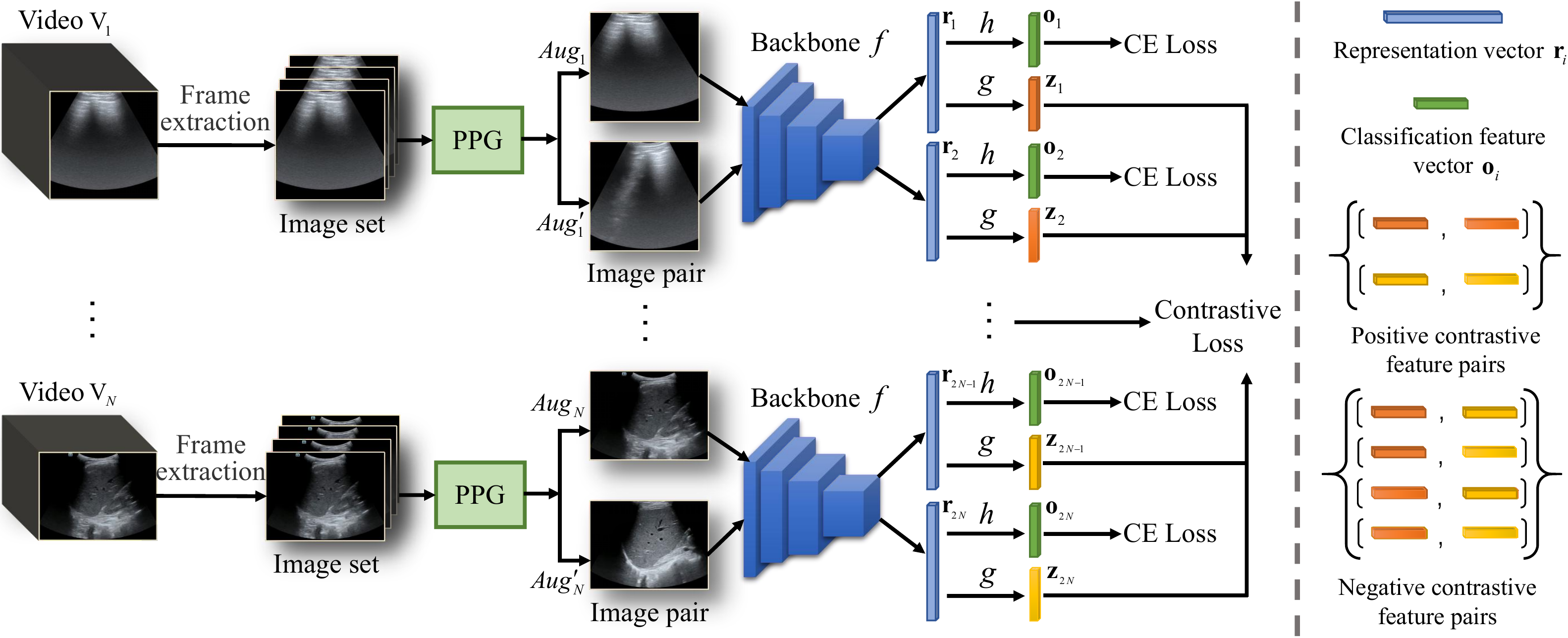} 
	\caption{System framework of the proposed USCL, which consists of sample pair generation and semi-supervised contrastive learning. (i) USCL extracts evenly distributed image sets from every US video as image dataset. (ii) The positive pair generation (PPG) module consists of a \textit{sampler} $\Theta$ random sampling several images from an image set, and a \textit{mixed frame generator} $G$ obtaining two images. A generated positive pair is processed by two separate data augmentation operations. (iii) A backbone $f$, a projection head $g$ and a classifier $h$ are trained simultaneously by minimizing the self-supervised contrastive learning loss and supervised CE loss.}
	\label{fig:framework}
\end{figure}

This section first formulates the proposed USCL framework (Fig. \ref{fig:framework}), then describes the details of sample pair generation. Finally, the proposed USCL will be introduced.

\subsection{Problem Formulation}
Given a video $V_i$ from the US-4 dataset, USCL first extracts images to obtain a balanced distributed frame set $\mathbb{F}_i^{K} \!=\!\{\textbf{f}_i^{(k)}\}^K_{k=1}$, where $K$ is the number of extracted images. Next, a \textit{sampler} $\Theta$ is applied to randomly sample $M$ images, denoted as $\mathbb{F}_i^{M}\!=\!\{\textbf{f}_i^{(m)}\}^M_{m=1}$ with $2 \leq M \ll K$. A following \textit{mixed frame generator} $G: \mathbb{F}_i^{M} \to \mathbb{S}_i^{2}$ obtains two images, where $\mathbb{S}_i^2\!=\!\{\textbf{x}^{(1)}_i, \textbf{x}^{(2)}_i \}$ is a positive pair followed by two data augmentation operations $Aug\!=\!\{{Aug}_i, {Aug'}_i\}$. These augmentations including random cropping, flipping, rotation and color jittering are used for perturbing positive pairs, making the trained backbones invariant to scale, rotation, and color style.

The objective of USCL is to train a backbone $f$ from training samples $\{(\textbf{x}_i^{(1)},\textbf{x}_i^{(2)}), \textbf{y}_i\}^N_{i=1}$ by combining self-supervised contrastive learning loss $\mathcal{L}_{con}$ and supervised cross-entropy (CE) loss $\mathcal{L}_{sup}$, where $N$ is the number of videos in a training batch. Therefore, the USCL framework formulation aims to minimize following loss $\mathcal{L}$:
\begin{equation}
\begin{split}
\mathcal{L} =  \mathcal{L}_{con}(g(f(Aug(G(\textbf{f}));\textbf{w}_f);\textbf{w}_g)) + \lambda \mathcal{L}_{sup}(h(f(Aug(G(\textbf{f}));\textbf{w}_f);\textbf{w}_h);\textbf{y}),
\label{eq:formulation}
\end{split}
\end{equation}
where $\lambda$ is a hyper-parameter, $\textbf{f}=\{\{\textbf{f}_i^{(m)}\}_{m=1}^{M}\}_{i=1}^N$ are frames sampled from a batch of videos for training, $G(\textbf{f})\!=\!\{\textbf{x}^{(1)}_i, \textbf{x}^{(2)}_i\}^N_{i=1}$ are positive pairs, and $\textbf{y} \!=\! \{\textbf{y}_i\}^N_{i=1}$ are corresponding class labels. $f$, $g$ and $h$ denote backbone, projection head (two-layer MLP) and linear classifier, respectively. Different from most existing contrastive learning methods, USCL treats contrastive loss in Eq. (\ref{eq:formulation}) as a consistency regularization (CR) term, which improves the performance of pretraining backbone by combining supervised loss in a mutually reinforcing way. Here, label information instructs the model to recognize samples with different labels to be negative pairs, and contrastive process learns how US images can be semantically similar or different to assist better classification.

\subsection{Sample Pair Generation}
\label{sect:FrameSelection}
Most of the existing contrastive learning approaches construct positive pairs by applying two random data augmentations image by image. When directly applying them to the US frames, the contrastive learning fails to work normally due to the similarity conflict problem (\ie two samples coming from the same video are too similar to be a negative pair). To solve this problem, a sample pair generation (SPG) scheme is designed: it generates positive pairs with the positive pair generation (PPG) module\footnote[7]{For more details of PPG module, see the Supplementary Material Section 2.}, and any two samples from different positive pairs are regarded as a negative pair. 

The PPG module regards an evenly distributed image set extracted from a video as a semantic cluster, and different videos belong to different clusters. This kind of organization fits the purpose of contrastive learning properly. We expect the model can map the semantic clusters to feature clusters. Then PPG generates two images as a positive sample pair from each cluster. Note that only one positive pair is generated from a video, which can prevent the aforementioned similarity conflict problem.

In detail, firstly, a \textit{sampler} $\Theta$ is applied to randomly sample three images $\widehat{\textbf{x}}_i^{(1)} , \widehat{\textbf{x}}_i^{(2)}$, and $\widehat{\textbf{x}}_i^{(3)}$ in chronological order from an image set $\{\textbf{f}_i^{(m)}\}_{m=1}^M $.
Secondly, a delicate \textit{mixed frame generator} $G$ is performed to generate a positive sample pair. The image $\widehat{\textbf{x}}_i^{(2)}$ is set as the anchor image, while $\widehat{\textbf{x}}_i^{(1)}$ and $\widehat{\textbf{x}}_i^{(3)}$ are perturbation images. In a mini-batch, $G$ constructs positive sample pairs in interpolation manner via the mixup operation between anchor image and two perturbation images as follows.
\begin{equation}
\left\{\begin{array}{l}
(\textbf{x}_i^{(1)}, \textbf{y}_i^{(1)}) = \xi_1 (\widehat{\textbf{x}}_i^{(2)}, \widehat{\textbf{y}}_i^{(2)}) + (1-\xi_1) (\widehat{\textbf{x}}_i^{(1)}, \widehat{\textbf{y}}_i^{(1)})\\
(\textbf{x}_i^{(2)}, \textbf{y}_i^{(2)}) = \xi_2 (\widehat{\textbf{x}}_i^{(2)}, \widehat{\textbf{y}}_i^{(2)}) + (1-\xi_2) (\widehat{\textbf{x}}_i^{(3)}, \widehat{\textbf{y}}_i^{(3)})
\end{array},\right.
\label{eq:mixup}
\end{equation}
where $\{\widehat{\textbf{y}}_i^{(k)}\}_{k=1}^3$ are corresponding labels. $\xi_1, \xi_2 \sim Beta(\alpha, \beta)$, where $\alpha, \beta$ are parameters of $Beta$ distribution. 

In our contrastive learning process, sample pairs are then fed to the backbone followed by the projection head for contrastive learning task. The proposed PPG module has several benefits: 1) Interpolation makes every point in the feature convex hull enclosed by the cluster boundary possible to be sampled, making the cluster cohesive as a whole; 2)  Positive pairs generated with Eq. (\ref{eq:mixup}) have appropriate mutual information. On the one hand, positive pairs are random offsets from the anchor image to the perturbation images, which ensures that they share the mutual information from the anchor image. On the other hand, the sampling interval $I \ge 5 $ frames in US-4, resulting in low probability for SPG to sample temporarily close $\{\widehat{\textbf{x}}_i^{(k)}\}_{k=1}^3$ which are too similar.

\subsection{Ultrasound Contrastive Learning}
\label{sect:ContrastiveLearning}
The proposed USCL method learns representations not only by the supervision of category labels, but also by maximizing/minimizing agreement between positive/negative pairs as CR. Here, assorted DNNs can be used as backbone $f$ to encode images, where the output representation vectors $ \textbf{r}_{2i-1}\!=\! f(\textbf{x}_i^{(1)})$ and $\textbf{r}_{2i} \!=\! f(\textbf{x}_i^{(2)})$ are then fed to the following projection head and classifier.

\myPara{Contrastive Branch.}
The contrastive branch consists of a projection head $g$ and corresponding contrastive loss.
The $g$ is a two layer MLP which nonlinearly maps representations to other feature space for calculating contrastive regularization loss. The mapped vector $ \textbf{z}_i\!=\!g(\textbf{r}_i) \!=\! \textbf{w}_g^{(2)}\sigma(\textbf{w}_g^{(1)}\textbf{r}_i) $ is specialized for a contrast, where $\sigma$ is ReLU activation function and $\textbf{w}_g\!=\!\{\textbf{w}_g^{(1)},\textbf{w}_g^{(2)}\}$ are the weights of $g$. 
The contrastive loss is proposed by Sohn~\cite{sohn2016improved}, which aims at minimizing the distance between positive pairs $\{ \textbf{x}_i^{(1)},\textbf{x}_i^{(2)} \}_{i=1}^N$ and maximizing the distance between any negative pair $\{\textbf{x}_i^{(1/2)},\textbf{x}_j^{(1/2)}\},~i \! \ne \!j$ for CR:
\begin{equation}
\mathcal{L}_{con} = \frac{1}{2N}\sum_{i=1}^{N}(l(2i-1,2i)+l(2i,2i-1)),
\label{eq:consistency}
\end{equation}
where
\begin{equation}
l(i,j) =-\log\frac{\exp(s_{i,j}/\tau)}{\sum_{k=1}^{2N}\mathbbm{1}_{i\ne k} \cdot exp(s_{i,k}/\tau)} \label{eq:ll},
\end{equation}
and
\begin{equation}
s_{i,j} = \textbf{z}_i \cdot \textbf{z}_j/(\|\textbf{z}_i\| \|\textbf{z}_j\|),
\label{eq:ss}
\end{equation}
where $\tau$ is a tuning temperature parameter. 

\myPara{Classification Branch.} We use a linear classifier $h$ with weights $\textbf{w}_c$ to separate sample features linearly in the representation space similar to \cite{he2016deep,huang2017densely}. The classification loss with corresponding one-hot label $\textbf{y}_i$ is
\begin{equation}
\mathcal{L}_{sup} = \frac{1}{2N}\sum_{i=1}^{N}(CE(\textbf{o}_{2i-1},\textbf{y}_{i})+CE(\textbf{o}_{2i},\textbf{y}_{i})),  
\label{eq:classification}
\end{equation}
where $\textbf{o}_i \!=\! h(\textbf{r}_i) \!=\! softmax(\textbf{w}_c \textbf{r}_i) $. 

Note that USCL is a semi-supervised training method, only contrastive branch works when the framework receives unlabeled data.
This semi-supervised design is intuitively simple but effective, which makes it easy to be implemented and has great potential to be applied to various pretraining scenarios.

\section{Experiment}
\subsection{Experimental Settings}
\myPara{Pretraining Details.}
ResNet18 is chosen as a representative backbone. We use US-4 dataset (the ratio of training set to validation set is 8 to 2) with 1\% labels for pretraining, and fine-tune pretrained models for various downstream tasks. During pretraining, US images are randomly cropped and resized to $224\!\times\!224$ pixels as the input, followed by random flipping and color jittering. We use Adam optimizer with learning rate $3\times10^{-4}$ and weight decay rate $10^{-4}$ to optimize network parameters. The backbones are pretrained on US-4 for $300$ epochs with batch size $N\!=\!32$. The pretraining loss is the sum of contrastive loss and standard cross-entropy loss for classification. Like SimCLR, the backbones are used for fine-tuning on target tasks, projection head $g$ and classifier $h$ are discarded when the pretraining is completed. The $\lambda$ in Eq.~(\ref{eq:formulation}) is $0.2$, parameters $\alpha$ and $\beta$ in Eq.~(\ref{eq:mixup}) are $0.5$ and $0.5$, respectively. The temperature parameter $\tau$ in Eq.~(\ref{eq:ll}) is $0.5$. The experiments are implemented using PyTorch with an Intel Xeon Silver 4210R CPU@2.4GHz and a single Nvidia Tesla V100 GPU.

\myPara{Fine-tuning Datasets.}
We fine-tuned the last 3 layers of pretrained backbones on POCUS~\cite{born2020pocovid} and UDIAT-B~\cite{2017Automated} datasets to testify the performance of our USCL. On POCUS and UDIAT-B datasets, the learning rates are 0.01 and 0.005, respectively. The POCUS is a widely used lung convex probe US dataset for COVID-19 consisting of 140 videos, 2116 images from three classes (\ie COVID-19, bacterial pneumonia and healthy controls). The UDIAT-B consists of 163 linear probe US breast images from different women with the mean image size of 760$\times$570 pixels, where each of the images presents one or more lesions. Within the 163 lesion images, 53 of them are cancerous masses and other 110 are benign lesions. In this work, we use UDIAT-B dataset to perform the lesion detection and segmentation comparison experiments. 50 of 163 images are used for validation and the rest are used for training.

\begin{table}[]
    \centering
    \caption{Ablation study of two contrastive ingredients during pretraining: assigning a negative pair from samples of different videos to overcome similarity conflict ($I_1$) and using mixup operation to enrich the features of positive pairs ($I_2$). They both improve the model transfer ability significantly, and the classification brunch is also beneficial. All results are reported as POCUS fine-tuning accuracy.}
    \begin{tabular}{c|c|c|c|cc|c|c|c|c|c}
				\hline	
				\multicolumn{1}{c|}{\multirow{3}{*}{Method}} & & \multicolumn{1}{c|}{\multirow{3}{*}{~ImageNet~}} & & \multicolumn{1}{c|}{\multirow{3}{*}{~USCL~}} &
			 $I_1$ & & \checkmark & & \checkmark & \checkmark \\
				\multicolumn{1}{c|}{} & & \multicolumn{1}{c|}{} & & \multicolumn{1}{c|}{} & $I_2$ & &  & \checkmark & \checkmark & \checkmark \\
				\multicolumn{1}{c|}{} & & \multicolumn{1}{c|}{} & & \multicolumn{1}{c|}{} & ~CE loss~ & & & & & \checkmark \\
				\hline
				Accuracy (\%)~~ & & ~84.2~ & & & & ~87.5~ & ~90.8~ & ~92.3~ & ~93.2~ & ~94.2~ \\ 
				\hline
    \end{tabular}
    \label{tab:ablation}
\end{table}

\begin{figure}[t]
\centering\centerline{\includegraphics[width=0.9\linewidth]{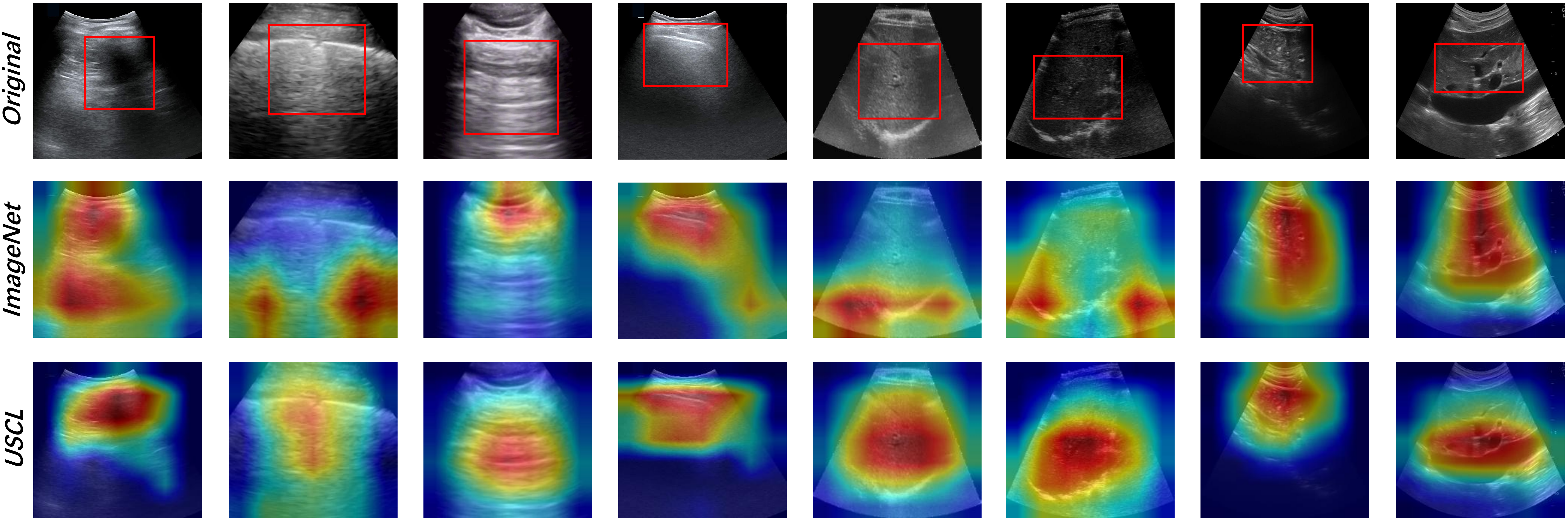}}
\caption{The visualization results of the last Conv layer in ImageNet pretrained model and USCL model with Grad-CAM~\cite{Selvaraju2017ICCV}. The first 4 columns are lung US images, models trained with USCL on US-4 can focus on the regions of A-line and pleural lesion instead of concentrating on regions without valid information like the ImageNet counterpart. The last 4 columns are liver US images, models trained with USCL accurately attend to the liver regions.}
\label{fig:Visualization}
\end{figure}

\subsection{Ablation Studies}
Here, we report the last 3 layers fine-tuning results on POCUS of US-4 pretrained backbones (ResNet18) to validate different components of USCL.

\myPara{SPG $\&$ CE loss.} We implement five pretraining methods considering the influence of different contrastive ingredients and the classification brunch with CE loss (Tab.~\ref{tab:ablation}). Compared with ImageNet, vanilla contrastive learning improves the accuracy by $3.3\%$ due to a smaller domain gap. It is regarded as the method baseline. The negative pair assigning scheme and positive pair generation method further improves the fine-tuning performance by 3.3\% and 4.8\%. They can be combined to reach higher performance. In addition, CE loss improves fine-tuning accuracy by $1.0\%$. This indicates that extra label information is able to enhance the power of contrastive representation learning.

\myPara{Visualization of Feature Representation.} To illustrate the robust feature representation of pretrained backbone, we visualize the last Conv feature map of some randomly selected images produced by USCL pretrained model and ImageNet pretrained model with Grad-CAM~\cite{Selvaraju2017ICCV} (Fig.~\ref{fig:Visualization}). Compared with ImageNet pretrained backbone, attention regions given by the USCL backbone are much more centralized and more consistent with clinical observation.

\subsection{Comparison with SOTA}
We compare USCL with ImageNet pretrained ResNet18~\cite{he2016deep} and other backbones pretrained on US-4 dataset with supervised method (\ie plain supervised), semi-supervised methods (\ie Temporal Ensembling (TE)~\cite{laine2016temporal}, $\Pi$ Model~\cite{laine2016temporal}, FixMatch~\cite{sohn2020fixmatch}), and self-supervised methods (\ie MoCo v2~\cite{chen2020improved}, SimCLR~\cite{chen2020simple}).

\myPara{Results on Classification Task.}
On POCUS dataset, we fine-tune the last three layers to testify the representation capability of backbones on classification task (Tab.~\ref{tab:POCUS_UDIAT}). USCL has consistent best performance on classification of all classes, and its total accuracy of $94.2\%$ is also significantly better than all 7 counterparts. Compared with ImageNet pretrained backbone, USCL reaches a much higher F1 score of 94.0\%, which is 12.2\% higher.

\myPara{Results on Detection and Segmentation Tasks.}
\label{detseg}
Tab.~\ref{tab:POCUS_UDIAT} shows the comparison results of detection and segmentation on UDIAT-B dataset. Mask R-CNN~\cite{kaiming2020TPAMI} with ResNet18-FPNs~\cite{Tsung2017CVPR}, whose backbones are pretrained, is used to implement this experiment. USCL generates better backbones than ImageNet and US-4 supervised learning. For detection and segmentation, it outperforms ImageNet pretraining by 4.8\% and 4.6\%, respectively. Importantly, the UDIAT-B images are collected with linear probe instead of convex probe like US-4, showing a superior texture encoding ability of USCL.
\begin{table}[t]
	\scriptsize
	\renewcommand\arraystretch{0.90}
	\caption{Comparison of fine-tuning accuracy ($\%$) on POCUS classification dataset and average precision (AP~\cite{lin2014microsoft})\protect\footnotemark on UDIAT-B detection (Det), segmentation (Seg) with SOTA methods.}
	\label{tab:POCUS_UDIAT}
	\setlength{\tabcolsep}{1.0mm}{
		\begin{center}
			\begin{tabular}{c|ccccc|c|c}
				\hline	\multicolumn{1}{c|}{\multirow{2}{*}{Method}} &\multicolumn{5}{c|}{{Classification}} &
			 \multicolumn{1}{c|}{~~~~~Det~~~~~} & \multicolumn{1}{c}{~~~~Seg~~~~~} \\
				\multicolumn{1}{c|}{} & COVID-19 & Pneumonia & Regular & Total Acc & ~~~~~F1~~~~ & AP & AP\\
				\hline
				ImageNet~\cite{he2016deep}  & 79.5  & 78.6  &  88.6 &  84.2   & 81.8 & 40.6  &  48.2  \\
				US-4 supervised   &  83.7 &  82.1 & 86.5  & 85.0 & 82.8 & 38.3 & 42.6\\
				\hline
				TE~\cite{laine2016temporal}  &   75.7 & 70.0 & 89.4  &    81.7 &  79.0 & 38.7 & 46.6\\
				$\Pi$ Model~\cite{laine2016temporal}    & 77.6 &  76.4 & 88.7 & 83.2  &  80.6 & 36.1 & 45.5\\
				FixMatch~\cite{sohn2020fixmatch}    &   83.0 & 77.5   &85.7  & 83.6   & 81.6 & 39.6 & 46.9 \\
				MoCo v2~\cite{chen2020improved}    &  79.7 & 81.4 & 88.9 & 84.8  & 82.8 & 38.7 & 47.1 \\
				SimCLR~\cite{chen2020simple}  & 83.2  &  89.4 &  87.1 & 86.4  & 86.3 &43.8 & 51.3\\
				\hline
				\textbf{USCL}    & \textbf{90.8} &  \textbf{97.0}   & \textbf{95.4}  &  \textbf{94.2} &  \textbf{94.0} &   \textbf{45.4} &  \textbf{52.8}\\
				\hline
			\end{tabular}
		\end{center}
	}
\end{table}
\footnotetext{AP is calculated as the area under the precision-recall curve drawn with different Intersection over Union (IoU) thresholds.}

\section{Conclusion}
This work constructs a new US video-based image dataset US-4 and proposes a simple but efficient contrastive semi-supervised learning algorithm USCL for US analysis model pretraining. USCL achieves significantly superior performance than ImageNet pretraining by learning compact semantic clusters from US videos. Future works include adding more scan regions of US videos to US-4 dataset for a better generalization on more diseases.

\section{Acknowledgement}
This work is supported by the Key-Area Research and Development Program of Guangdong Province (2020B0101350001); the GuangDong Basic and Applied Basic Research Foundation (No. 2020A1515110376); Guangdong Provincial Key Laboratory of Big Data Computation Theories and Methods, The Chinese University of Hong Kong (Shenzhen).

%
\bibliographystyle{splncs04}
\bibliography{USCLbib}
\end{document}